%% file: main.tex
\documentclass[sigconf]{acmart}
\makeatletter
\def\@ACM@checkaffil{% Only warnings
    \if@ACM@instpresent\else
    \ClassWarningNoLine{\@classname}{No institution present for an affiliation}%
    \fi
    \if@ACM@citypresent\else
    \ClassWarningNoLine{\@classname}{No city present for an affiliation}%
    \fi
    \if@ACM@countrypresent\else
        \ClassWarningNoLine{\@classname}{No country present for an affiliation}%
    \fi
}
\makeatother
\usepackage{fancyhdr}

\renewcommand\footnotetextcopyrightpermission[1]{} % removes footnote with conference information in first column
\AtBeginDocument{%
  \providecommand\BibTeX{{%
    \normalfont B\kern-0.5em{\scshape i\kern-0.25em b}\kern-0.8em\TeX}}}

\usepackage{multirow}
\usepackage{graphicx}
\usepackage{subfigure}
\usepackage{threeparttable}
\usepackage{hyperref}
\usepackage{url}
\usepackage{color}

\usepackage{enumitem}
\usepackage{booktabs}
\usepackage[skip=5pt]{caption}
\usepackage[normalem]{ulem}

\def\geolm{\textsc{K2}}

\begin{document}
\begin{sloppypar}
\title{
% \geolm{}: The Leading Foundation Model for Geoscience 
% Knowledge Understanding
% Learning Foundation Language Models for Geoscience Knowledge Understanding and Utilization
\geolm{}: A Foundation Language Model for Geoscience Knowledge Understanding and Utilization
}

\author{Cheng Deng$^1$, Tianhang Zhang$^1$, Zhongmou He$^1$, Yi Xu$^1$, Qiyuan Chen$^2$, Yuanyuan Shi$^1$, \\Luoyi Fu$^1$, Weinan Zhang$^1$, Xinbing Wang$^1$, Chenghu Zhou$^3$, Zhouhan Lin$^1$, Junxian He$^1$}
\affiliation{%
  \institution{$^1$Shanghai Jiao Tong University, $^2$University of Waterloo}
  \institution{$^3$Institute of Geographical Science and Natural Resources Research, Chinese Academy of Sciences}
}
% \email{{davendw, zhangtianhang}@sjtu.edu.cn, zhouch@lreis.ac.cn}
\affiliation{Corresponding authors: Zhouhan Lin and Junxian He}

\renewcommand{\shortauthors}{Cheng Deng, et al.}

\begin{abstract}
  Large language models (LLMs)
  % , acting like foundation models, 
  have achieved great success in general domains of natural language processing.
  In this paper, we bring LLMs to the realm of geoscience with the objective of advancing research and applications in this field. 
  % while the application over specialized domains such as geoscience remains a challenge, creating barriers to improving the working efficiency of geoscientists.
  To this end, we present the first-ever LLM in geoscience, \textbf{\geolm{}}, alongside a suite of resources developed to further promote LLM research within geoscience. For instance, we have curated the first geoscience instruction tuning dataset, \textbf{GeoSignal}, which aims to align LLM responses to geoscience-related user queries. Additionally, we have established the first geoscience benchmark, \textbf{GeoBench}
  % \jh{I suggest to rename this benchmark as ``GeoBench'' throughout the paper.}
  , to evaluate LLMs in the context of geoscience.
  % we introduce \textbf{\geolm{}}, currently the first LLM in geoscience, along with a set of resources to build it to promote futher LLM research on geoscience. 
  % For example, we curate the first geoscience instruction tuning dataset, \textbf{GeoSignal}, to align LLMs follow geoscience-related user queries. Moreover, we build the first geoscience benchmark, \textbf{GeoBench}, to evaluate geoscience LLMs. 
  In this work, we experiment with a complete recipe to adapt a pre-trained general-domain LLM to the geoscience domain. 
  Specifically, we further train the LLaMA-7B model on 5.5B tokens of geoscience text corpus,
  % \jh{Are the 5.5B tokens domain-specific? If so, please specify}
  including over 1 million pieces of geoscience literature, and utilize GeoSignal's supervised data to fine-tune the model. Moreover, we share a protocol that can efficiently gather domain-specific data and construct domain-supervised data, even in situations where manpower is scarce.
  Meanwhile, we equip \geolm{} with the abilities of using tools to be a naive geoscience aide.
  % Specifically, we further pretrain the LLaMA-7B model over 1 million geoscience literature and use GeoSignal supervised data to finetune the model. 
  % We also share a recipe that can efficiently collect domain data and build domain-supervised data in the absence of manpower. 
  Experiments conducted on the GeoBench demonstrate the effectiveness of our approach and datasets on geoscience knowledge understanding and utilization.
  % Moreover, we evaluate our model over GeoBench cconsisting the AP Test of Geology and the Chinese Postgraduate Entrance Examination on Geoscience. Finally, the result shows the domain adaptation of geoscience. 
  We open-source all the training data and \geolm{} model checkpoints at \url{https://github.com/davendw49/k2}.
\end{abstract}





\maketitle

\input{intro}
\input{related}
\input{rst}
\input{train}

\input{evaluation}
\input{discussion}
\input{conclusion}
\newpage
\bibliographystyle{ACM-Reference-Format}
\bibliography{ref}

% \appendix
% \input{appendix}

\end{sloppypar}
\end{document}

%% file: intro.tex
\section{Introduction}
Geoscience, an interdisciplinary research field, is an integral subject in natural science, investigating the formation and evolution of the Earth~\cite{TheGeoImpact}. Geoscientists have long faced challenges in integrating data from various sources and disciplines due to differences in terminologies, formats, and data structures, which subsequently leads to number of natural language tasks in geoscience such as geological and geographical named entity recognition~\cite{Enkhsaikhan2021AutolabellingEI}, spatial and temporal relation extraction~\cite{Ma2020ANS} to build geoscience knowledge graph~\cite{Deng2021GAKGAM}, geology reports and literature summarization~\cite{Ma2021WhatIT}, and representation learning via geoscience language models~\cite{Padarian2019WordEF}. 
However, language models in geoscience are sparse and remain limited in scale~\cite{Denli2021GeoscienceLP}.
% \jh{add citation} 
This situation stands in stark contrast with the prosperity of large language models (LLMs), such as ChatGPT~\cite{openaichatgpt} and GPT-4~\cite{OpenAI2023GPT4TR}, in general natural language processing (NLP), where notable successes have been achieved.

Despite their effectiveness in general domains, current LLMs often fall short in catering to the needs of geoscientists. This shortfall is largely attributed to the lack of reliable knowledge concerning geoscience problems, given that the related geoscience data seldom exist in the commonly used pre-training text corpora such as C4~\cite{Raffel2019ExploringTL} and the Pile~\cite{Gao2020ThePA}. 
% so current large language models fail to be distilled out enough geoscience domain knowledge. Consequently, these issues have brought certain obstacles to the combination of natural language technology with geoscience.
Moreover, top-performing LLMs like ChatGPT only offer services via APIs, which presents roadblocks to external domain research and advancement.
To mitigate these issues and foster research and application within the geoscience domain, 
we introduce the first-ever open-source LLM for geoscience, referred to as \textbf{\geolm{}} (\textit{The second highest mountain in the world, where we believe in the future larger and more powerful geoscience language models will be created}).
\geolm{}, a GPT-like language model comprising 7 billion parameters, is based on the pre-trained LLaMA~\cite{Touvron2023LLaMAOA}
model but specializes in the geoscience domain. 
Along with the introduction of \geolm{}, this paper also explores a roadway to collect geoscience text corpus, constructs geoscience instruction supervised data, and builds geoscience NLP benchmarks, in alignment with the Deep-time Digital Earth (DDE,~\cite{Wang2021TheDD})\footnote{\href{https://www.iugs.org/dde}{https://www.iugs.org/dde}} big science plan.

\begin{figure*}[!t]
    \centering
    \includegraphics[width=0.9\linewidth]{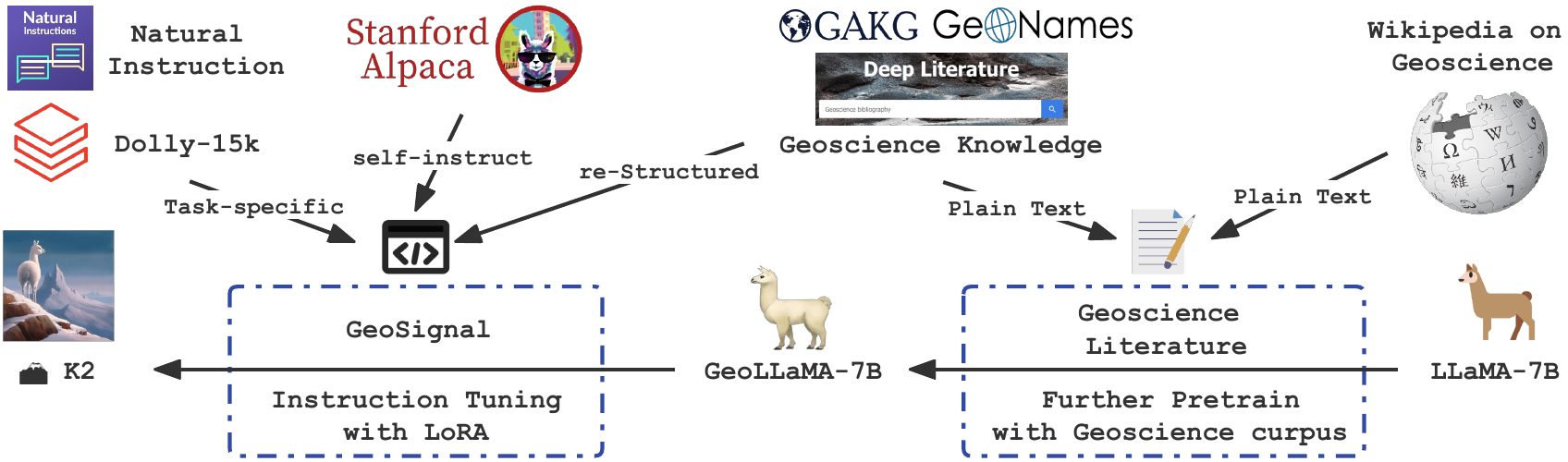}
    \caption{Pipeline of training \geolm{}, including two steps, one is further pre-train for absorption of geoscience knowledge, another one is instruction tuning, deploying to make the model align to human, instructed by human, and response like a human.}
    % \jh{images too vague, please replace with higher-resolution ones..}
    \label{fig:1}
    % \vspace{-1em}
\end{figure*}

The training of K2 consists of two stages, the pre-training stage and the instruction tuning stage, as depicted in \autoref{fig:1}. 
During pre-training, we continue pre-training the LLaMA-7B model on a geoscience text corpus that we preprocessed from geoscience papers. 
Then we perform instruction tuning~\cite{Sanh2021MultitaskPT,Longpre2023TheFC,Chung2022ScalingIL},
% ~\jh{cite multi-prompt tuning, flan, flan-t5} 
where we further train the model to follow human instructions. 
To this end, we have curated \textbf{GeoSignal}, an instruction tuning dataset created by unifying the examples from 8
% \jh{20?}
diverse geoscience NLP tasks with prompts, such as relation extraction, entity recognition, classification, and summarization.
We also construct \textbf{GeoBench}, an evaluation dataset comprising more than 1500 objective questions and 939 subjective questions collected from National Postgraduate Entrance Examination (NPEE) on Geoscience and AP Test Geology, Geography, and Environmental Science.
GeoBench serves to track the progress and drive the development of geoscience language models.
% Our primary focus is to enhance an open-source large language model, LLaMA, with knowledge of geoscience and better understand and utilize the field. By further pre-training LLaMA with geoscience text corpus and finetuning on our re-structured geoscience domain supervised instructions, we finally create the first foundation large language model in geoscience, named \textbf{\geolm{}} (\textit{The second highest mountain in the world, which we believe in the future the larger and the more powerful geoscience language model will be created}) and evaluate the \geolm{} on newly constructed geoscience NLP tasks benchmarks.
Through our concerted efforts in data collection and training, the resulted \geolm{} model is a foundation language model that can be used to design multiple geoscience applications, making it benefit geoscience researchers and practitioners~\cite{Mai2023OnTO}. 
% The pipeline for training \geolm{} is shown in \autoref{fig:1}.
To exemplify this, we train \geolm{} to learn to use geoscience academic search tools through tool learning and, simultaneously, guide \geolm{} to do the relation prediction between geoscience-related knowledge points
% \jh{What is ``connection between geoscience-related knowledge points''? is this like relation prediction? Please clarify.} 
through chain-of-thought\cite{Wei2022ChainOT} and generate new ideas. Therefore, \geolm{} demonstrates its potential in geoscience knowledge mining and research assistants.

Our contributions can be listed as follows:
\begin{itemize}[leftmargin=0.8em]
    \item We introduce \geolm{}, a foundation langugage model in geoscience field. \geolm{} can answer geoscience questions, follow geoscientists' instructions via suitable prompts with its professionalism in geoscience, and have the ability to use tools as extensions.
    \item We construct GeoSignal, the first-ever geoscience-supervised instruction data. To evaluate \geolm{} on geoscience tasks and the following language models in geoscience, we build GeoBench, the first NLP task benchmarks in geoscience.
    \item Taking geoscience as an example, we build up a paradigm to construct the domain text corpus, domain-supervised instruction data and explore a recipe to train a domain-specific LLM.
    \item Compared with similar-size baseline models, \geolm{} outperforms both subjective and objective geoscience tasks, including taking examinations and doing knowledge reasoning.
    % \jh{if this is single-blind review, I suggest we directly release all the materials near the submission time}
    At last, we release all the code, \geolm{} weights, GeoSignal, and GeoBench at \href{https://github.com/davendw49/k2}{https://github.com/davendw49/k2}.
    % \jh{we release the code, \geolm{} weights, GeoSignal, and GeoBenchmark at xxx}
\end{itemize}
The rest of the paper is arranged as follows: 
Section 2 will introduce the related work of \geolm{}. 
In Section 3, the detail of data collection, supervised instruction data construction, and benchmarks construction will be illustrated.
Further, we will share our further pre-training details and parameter-efficient instruction tuning processes in Section 4. In Section 5, we will evaluate the \geolm{} and perform ablation studies.
Finally, in Section 6, we will share the application of \geolm{}, showing that \geolm{} has great potential for geoscience scientific utilization.

%% file: related.tex
\section{Related Work}
\textbf{Foundation Language Models.} Since the appearance of ChatGPT~\cite{openaichatgpt}, there has been a large number of large language models for use as a foundation model to solve real-life problems. Since the models that provide only online demos and APIs, like ChatGPT, GPT-4\cite{OpenAI2023GPT4TR} and Yiyan (\url{https://yiyan.baidu.com/}) are not suitable and convenient for further pre-training and developing. The open-source models like CodeGen~\cite{Nijkamp2022CodeGenAO}, LLaMA\cite{Touvron2023LLaMAOA}, GLM~\cite{Zeng2022GLM130BAO}, 
% GPT-2~\cite{radford2019language}, GPT-J~\cite{gpt-j}, T5~\cite{Raffel2019ExploringTL}, Cerebras-GPT~\cite{Dey2023CerebrasGPTOC}, Pythia~\cite{Bierlich2022ACG},% and Bloom~\cite{Scao2022BLOOMA1} 
becomes the foundation models for many other instruction-tuned LLMs like Alpaca~\cite{alpaca}, Baize~\cite{Xu2023BaizeAO}, Vicuna~\cite{vicuna2023}, Koala~\cite{koala_blogpost_2023}, and Dolly~\cite{dolly}.
% Flan-T5~\cite{Chung2022ScalingIL} and .
% However, the demand for model updates is high, keep training LLMs are not the most appropriate way. 
Meanwhile, The tool learning has been brought to the fore by LLM as the rapid rise of the usage of external modules and information to enhance the foundation models. 

\noindent\textbf{Domain Language Models.} Large language models become the foundation model to address the issues in many other domains. 
In life science field, Med-PaLm~\cite{Singhal2022LargeLM}, MedGPT~\cite{Kraljevic2021MedGPTMC}, BioGPT~\cite{Luo2022BioGPTGP}, and Bio-Megatron~\cite{Shin2020BioMegatronLB}
%, BioBERT~\cite{Lee2019BioBERTAP} , ClincalBERT~\cite{Huang2019ClinicalBERTMC}, PubMedBERT~\cite{Gu2020DomainSpecificLM}, BioBART~\cite{Yuan2022BioBARTPA},  and , 
The large language model is useful and reliable in the biomedicine field~\cite{Wang2021pretrainedLM}.
In natural science field, Geographic-BERT~\cite{Liu2021GeoBERTPM}, MGeo~\cite{Ding2023AMG}, PK-Chat~\cite{deng2023pk}, ERNIE-GeoL~\cite{Huang2022ERNIEGeoLAG} and GeoBERT~\cite{Denli2021GeoscienceLP} are typical cases in geography and geology~\cite{Mai2021ARO}, while MatSciBERT~\cite{Gupta2021MatSciBERTAM} is the one in material science. 
% Moreover, LEGAL-BERT~\cite{Chalkidis2020LEGALBERTTM}, BERTweet~\cite{Nguyen2020BERTweetAP}, FinBERT~\cite{Araci2019FinBERTFS}, COVID-twitter-BERT~\cite{Mller2020COVIDTwitterBERTAN} shows the power of language model can reason over social media and business data.
In academic scenario, SciBERT~\cite{Beltagy2019SciBERTAP} and Galactica~\cite{Taylor2022GalacticaAL} are two examples. 

% \noindent\textbf{Data Engineering} \jh{you can remove this paragraph if you are short of space}Currently, high-quality text corpus and supervised instruction data are treasures. The fundamental text corpus for pre-train stage are mainly based on the Pile~\cite{Gao2020ThePA}, C4~\cite{Raffel2019ExploringTL}, Common Crawl~\cite{commoncrawl}, Wikipedia, CLUECorpus~\cite{CLUECorpus2020}, Public Git Archive~\cite{Markovtsev2018PublicGA} and ArXiv using in Galactica.
% Regarding instruction data, the Flan Collection~\cite{Longpre2023TheFC} is a general domain instruction data collection, but while facing the geoscience field, the lack of domain text corpus and instructions remains an issue. Self-instruct~\cite{Wang2022SelfInstructAL} sheds light on collecting data from language models themselves, and Baize generates the instruction via self-chatting between two ChatGPT bots. RST~\cite{Yuan2022reStructuredP} shares a pipeline to extract pair data from signal data, while Natural Instruction~\cite{mishra2021natural} data provide the scientific language models and academic instruction tuning data.

\noindent\textbf{Parameter-Efficient Tuning on LLMs.} Conventional fine-tuning needs to update all the parameters in LLMs, leading to inefficient and leaving a large carbon footprint as the models grow along with the scaling law~\cite{Kaplan2020ScalingLF}. Soft Prompt tuning~\cite{Lester2021ThePO} frozen language models to perform specific downstream tasks. Prefix-tuning~\cite{Li2021PrefixTuningOC} draws inspiration from prompting for language models, allowing subsequent tokens to attend to this prefix as if it were ``virtual tokens''. In addition, Adapter~\cite{Houlsby2019ParameterEfficientTL} make the parameters of the original network remain fixed, yielding a high degree of parameter sharing, and LoRA~\cite{Hu2021LoRALA} views the update of the weights as the result of two tunable low-rank matrices multiplication. 

%% file: rst.tex
\section{Data Collection and Curation}

To train \geolm{}, we collect geoscience text corpus and geoscience-oriented data from various resources. Then, we re-structure the data into signals and build up the instruction tuning dataset GeoSignal. This valuable information can serve for learning knowledge for geoscience tasks and instruct models for aligning with humans and experts. Moreover, we develop GeoBench to compare language models focusing on geoscience.
% We will publicly make our datasets available after the final draft on \href{https://github.com/davendw49/k2}{Github}.

\begin{figure*}[!t]
    \centering
    \includegraphics[width=\linewidth]{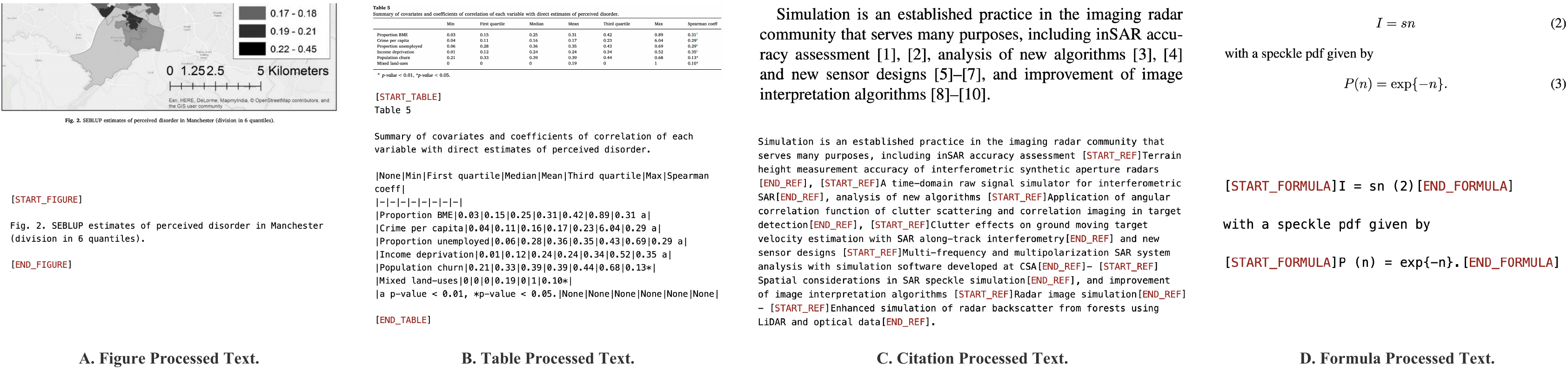}
    \caption{\textbf{Tokenization processed text.} \textbf{A.} shows an example of a figure marker, we only choose to preserve the captions; \textbf{B.} shows an example of a table marker, we transfer the tables into the form of Markdown; \textbf{C.} shows the tokenization of the citations, we replace the reference numbers into reference papers' title to preserve the readability of the text corpus; \textbf{D.} shows an example of the special tokens for formulas.}
    \label{fig:prepro}
\end{figure*}

\subsection{Pre-training Data}

In this work, our text corpus for further pre-training on LLaMA-7B consists of \emph{5.5 billion tokens}, including geoscience-related Wikipedia pages, geoscience paper's abstracts, and open-access geoscience papers published in selected high-quality journals in earth science and mainly collected by GAKG~\cite{Deng2021GAKGAM}. 

\subsubsection{Geoscience Text Corpus Collection} 

% \usepackage{booktabs}
\begin{table}[!t]
\begin{tabular}{@{}lrc@{}}
\toprule
\textbf{Data source}        & \textbf{Document} & \textbf{Tokens} \\ \midrule
Geoscience papers           & 1,122,094       &    3.9B  \\
Geoscience papers Metadata  & 4,274,716       &    0.1B  \\
Wikipedia pages             & 767,341         &    1.5B  \\
\textbf{Total}              & \textbf{6,164,151}  & \textbf{5.5B}      \\ \bottomrule
\end{tabular}
\caption{The details of the text corpus used to train \geolm{}.}
\label{tab:textcorpus}
\vspace{-1em}
\end{table}

\paragraph{\textbf{Geoscience Open Access Literatures.}} With the support from \href{https://deep-time.org/}{DDE} (Deep-Time Digital Earth Big Science Program), we can have the resources and chances to access materials and data strongly related to geoscience, including \emph{531} journals and \emph{4,274,716} papers' metadata. We use \emph{1,122,094} open-access papers' PDFs organized by GAKG\footnote{\href{https://gakg.acemap.info}{https://gakg.acemap.info}} to build the text corpus.
\paragraph{\textbf{Wikipedia pages about Earth science}} Wikipedia is an import resource we take into account for text corpus collection, and the root node of the Wikipedia category of geoscience we take into consideration is \textit{``https://en.wikipedia.org/wiki/Earth\_science''}. We mine all the child and related topics connected to it and finally gain \emph{767,341} Wikipedia pages. 

In brief, the statistics of the collection of geoscience text corpus are shown in~\autoref{tab:textcorpus}.
\subsubsection{Text Corpus Preprocessing} 

\paragraph{\textbf{PDF Parsing}}
We build an automatic PDF parsing toolkit based on the GROBID library~\cite{GROBID}. We use Markdown as the format for all papers in the corpus to preserve readability and consistency. Finally, we use regular expressions and rule-based scripts to clean the data, removing the text obstructing reading, garbled, and impurity data. The script will be released after the final draft, and currently, it is in use by DeepShovel~\cite{Zhang2022DeepShovelAO}.

\paragraph{\textbf{Tokenization}}
Tokenization is an essential part of text corpus design. To make the language model understand the academic papers, we utilize specialized tokens for different modalities as follows, and the examples are shown in~\autoref{fig:prepro}.
\begin{itemize}[leftmargin=0.8em]
    \item \textbf{Illustrations:} we use special tokens \textsc{[START\_FIGURE]} and \textsc{[END\_FIGURE]} to annotate the captions of the illustrations in the papers.
    \item \textbf{Tables:} Two special tokens \textsc{[START\_TABLE]} and \textsc{[END\_TABLE]} are used to locate the position of the table in the passage. In this process, we transform the tables in the PDFs into the format of Markdown.
    \item \textbf{Citations:} We use special tokens \textsc{[START\_REF]} and \textsc{[END\_REF]} to annotate the citations.
    \item \textbf{Formulas:} For mathematical content or formulas, we filter and clean the irregular formulas parsed from PDFs through regular expressions and rule-based methods. Further we use special tokens \textsc{[START\_FORMULA]} and \textsc{[END\_FORMULA]} to capture them.
\end{itemize}

\subsection{Instruction Tuning Data: GeoSignal}

Next, we curate the instruction tuning data that will be used to align the pre-trained model with user intentions.
Specifically, we first collect well-organized, general instruction tuning data, such as natural instruction~\cite{mishra2021natural}, AI2 Reasoning Challenge~\cite{Clark2018ThinkYH}, stanford-alpaca~\cite{alpaca}, and Dolly-15k~\cite{dolly}. Then, we utilize a semi-manual pipeline to build up a geoscience expert-alignment dataset called \textbf{GeoSignal}. Moreover, we create a tool training dataset based on ToolBench~\cite{Qin2023ToolLLMFL} to enable \geolm{} to use tools. These instruction tuning data statistics are shown in~\autoref{tab:sftdata}. We detail the data curation process next. 

% \usepackage{booktabs}
\begin{table}[!t]
\resizebox{\linewidth}{!}{%
\begin{tabular}{@{}lrl@{}}
\toprule
\textbf{SFT Data} & \textbf{Prompts}  & \textbf{Data Type}\\ \midrule
Alpaca-GPT4             &  52,002  & Self-instruct        \\
Dolly-15K               &  15,011  & Task-specific        \\
Natural Instruction     &  2,446   & Task-specific         \\
AI2 Reasoning Challenge &  7,787   & Task-specific         \\
GeoTool                 &  10,645  & Tools                 \\ 
\textbf{GeoSignal}      &  39,749  &\textbf{Knowledge Intensive}   \\ 
\bottomrule
\end{tabular}
}
\caption{Datasets used to train \geolm{} during the instruction tuning process.}
% \jh{Is the tool learning dataset exactly Toolbench or different from Toolbench? Your text implies they are different. If different, you should use a different name in the table. If you are just using Toolbench, you need to revise your text.}
\vspace{-2em}
\label{tab:sftdata}
\end{table}

\subsubsection{Align-to-Human} 

In this part, we collect and gather several well-construct supervised datasets, including self-instruct, human-annotated, and tools-related data. 

\begin{itemize}[leftmargin=0.8em]
    \item \textbf{Alpaca-GPT4:} Alpaca-GPT4\footnote{\url{https://github.com/tloen/alpaca-lora}} is an instruction-following dataset generated by the techniques named Self-Instruct~\cite{Wang2022SelfInstructAL}, and all the samples are in the form of \textit{<instruction, input, output>}, which we choose to follow.
    \item \textbf{Dolly-15k:} databricks-dolly-15k~\cite{dolly} is an open-source dataset of instruction-following records generated by thousands of Databricks employees, including brainstorming, classification, closed QA, generation, information extraction, open QA, and summarization. We organize them all into \textit{<instruction, input, output>} format.
    \item \textbf{Natural Instruction:} Natural Instruction~\cite{mishra2021natural} maintains many tasks and their natural language definitions/instructions. Its v1.x dataset consists of 61 tasks. The v2.x dataset contains over 1.5k tasks. We select \emph{objective tasks} elaborately from the v2.x dataset and organize them into \textit{<instruction, input, output>} format.
    \item \textbf{AI2 Reasoning Challenge:} AI2 Reasoning Challenge (ARC)~\cite{Clark2018ThinkYH} is a dataset of \emph{7,787} genuine grade-school level, multiple-choice science questions. As it is well-formed, we sample randomly and organize it into \textit{<instruction, input, output>} format.
    \item \textbf{Tool Instruction Data:} Refer to ToolBench~\cite{Qin2023ToolLLMFL}, We manually curated a collection of 2k instruction data for training \geolm{} to learn to use geoscience academic search engine (we name this tool as \emph{GeoSearch}).
    % \jh{can you add one sentence to be more specific here? for example, are the 2k data generated from ChatGPT? (because toolbench distills from chatGPT)}
    Then, we combine ~8k tool instruction data (\emph{arxiv, bing search, database, weather, and wolfram-alpha}) from ToolBench to prepare for the tool learning. We call this collection of tool instruction data \emph{GeoTool}, and it has 10k samples in total.
    % \jh{What is the GeoSearch data, have you described them yet?} 
    % \jh{Why do you say ``based on Toolbench''? Have you used Toolbench data actually? If not, you can just say ``Inspired by Toolbench'' or ``Similar to Toolbench''.}\jh{It will help if you can showcase an example of tool instruction data here.}
\end{itemize}

\subsubsection{Align-to-Expert} 

An expert is a human who specializes in a given domain, and more than learning to follow human instruction is needed for the specialized domain, we set up to train the model with knowledge-intensive data. Referring to re-structured pre-training~\cite{Yuan2022reStructuredP,Chung2022ScalingIL}, signals are the data we can use to train models and usually exist in databases and websites. Many data sources and materials have different types of geoscience signals in geoscience, as illustrated in \autoref{fig:geosignal}. 

These signals could be re-structured into \textit{<input, output>} pairs as instruction tuning samples. For example, with a paper's abstract and title information, we can re-structure such signals into a title generation task given the abstract. In addition, and most importantly, with the support of several applications and products of DDE, we collect a large quantity of geoscience expertise data and re-structure it with prompts into a unified sequence-to-sequence format, namely GeoSignal. The databases and websites we use are as follows:

\begin{figure*}[!t]
    \centering
    \includegraphics[width=0.95\linewidth]{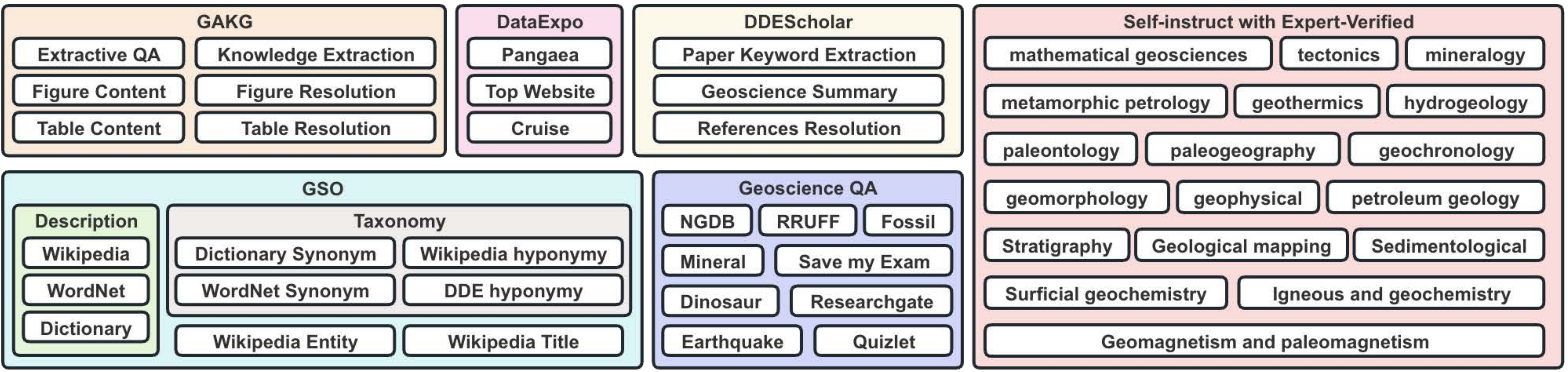}
    \caption{The components of GeoSignal.}
    \label{fig:geosignal}
    % \vspace{-1.3em}
\end{figure*}

\begin{itemize}[leftmargin=0.8em]
    \item \textbf{GAKG:} GAKG~\cite{Deng2021GAKGAM} is a multimodal Geoscience Academic Knowledge Graph organizing geoscience papers' illustrations, text, and bibliometric data. Each paper has several \textit{geor:mention\_knowledge} axioms to connect different knowledge points.
    \item \textbf{DDE Scholar:} DDE Scholar (\url{https://ddescholar.acemap.info/}), a geoscience academic literature search engine, contains more than 3 million papers and 4 million scholars' information in the field of earth sciences.
    \item \textbf{DataExpo:} DataExpo~\cite{Lu2023DataExpoAO} is a one-stop dataset service and has indexed over 960,000 datasets from more than 27,000 repositories in the context of Deep-time Digital Earth Program.
    \item \textbf{GSO:} GSO (\url{https://gso.acemap.info/}) is a large-scale ontology of research areas that was automatically generated using the hierarchical topic modeling, which consists of more than 120 thousand research interests in the field of geoscience.
    \item \textbf{Geoscience QA:} We crawler \emph{4} question and answer platform, and \emph{6} geoscience-related databases, using OpenAI~\cite{openaichatgpt} for template generation and with the help of the human expert, we finally have a clean and correct geoscience Q\&A dataset. The distribution of each part is shown in~\autoref{tab:geoqa}.
    \item \textbf{Self-instruct:} Refer to Alpaca-GPT4, we use GPT4 to generate 18,000 questions and their answers from 18 subfields of geoscience based on domain materials and verify most of the Q\&A pairs by geoscientists.~\footnote{In the data curation process and experiments throughout this paper, we use the 2023 March version of ChatGPT and 2023 March version of GPT-4 unless otherwise specified.}
\end{itemize}

% % Please add the following required packages to your document preamble:
% % \usepackage{booktabs}
% \begin{table}[h]
% \begin{tabular}{@{}llll@{}}
% \toprule
% \textbf{Resource} & \textbf{Count} & \textbf{Resource}  & \textbf{Count} \\ \midrule
% NGDB              & 148,212        & Earthquake         & 37,284          \\
% RRUFF             & 32,778         & SaveMyExam         & 1,107           \\
% Fossil            & 4,959          & ResearchGate       & 3,680           \\
% MinDat            & 51,291         & Quizlet            & 301             \\
% Dinosaur          & 11,348         & Study              & 1,294           \\ \midrule
% \multicolumn{2}{l}{\textbf{Total}} & \multicolumn{2}{r}{\textbf{292,254}} \\ \bottomrule
% \end{tabular}
% \caption{The statistics of the Geoscience QA data collection.}
% \label{tab:geoqa}
% \end{table}

For a better understanding of geoscience signals, we list the main signals we consider in bellowing and illustrate the detail of re-structure.

\begin{itemize}[leftmargin=0.8em]
    \item \textbf{G1: Paper content:} The title, abstract, full-text of geoscience literature. This signal naturally exists on DDE Scholar, GAKG, and DataExpo and can be used in summarization tasks.
    \item \textbf{G2: Category:} The category of a geoscience paper or term. This signal typically exists on DDE Scholar, GAKG, and Wikipedia. It can be used for the text classification task.
    \item \textbf{G3: Reference Paper:} This signal exists in the reference lists and introduction of papers and is useful for text comprehension and summarization.
    \item \textbf{G4: The captions of paper table and illustration:} Tables and figures in geoscience papers provide captions and content mentioned in the passage, which can be used for question-answering tasks.
    \item \textbf{G5: Entity mentions:} The entities within a given text. This signal can be found in GAKG and Wikipedia and can be useful for named entity recognition tasks.
    \item \textbf{G6: Relations:} The relationships between different geoscience entities. This information exists in human-annotated datasets such as GAKG and GSO. This signal is useful for finding synonyms and hyponymy terms in geoscience.
    \item \textbf{G7: Word description:} The definition of a word. Various geoscience resources contain this signal, such as Geoscience Dictionary, WordNet, Wikipedia, and GSO. This signal is useful for the task of explanation.
    \item \textbf{G8: Synonyms \& Taxonomy:} The Synonyms and hyponymy relation between terms in geoscience. Geoscience Dictionary and GSO contain this signal, useful for finding synonyms and hyponymy terms in geoscience.
    \item \textbf{G9: Text Comprehension:} This signal typically exists in geoscience academic platforms and other text material containing question and answer pairs and is useful for question answering.
    \item \textbf{G10: Factual knowledge:} Geoscience facts, e.g., Dolomite is a carbonate rock. This signal typically exists in some geoscience-related QA platforms and is useful for question-answering and fact verification.
\end{itemize}

% \usepackage{booktabs}
\begin{table}[!t]
\resizebox{\linewidth}{!}{%
\begin{tabular}{@{}lll@{}}
\toprule
\textbf{Resource} & \textbf{Count}   & \textbf{Link}                                    \\ \midrule
NGDB              & 148,212          & \url{https://mrdata.usgs.gov/}                         \\
RRUFF             & 32,778           & \url{https://rruff.info/}                              \\
Fossil            & 4,959            & \url{http://fossil-ontology.com/}                      \\
MinDat            & 51,291           & \url{https://zh.mindat.org/}                           \\
Dinosaur          & 11,348           & \url{https://dinoanimals.com/dinosaurdatabase/}        \\
Earthquake        & 37,284           & \url{https://public.opendatasoft.com/}                 \\
SaveMyExam        & 1,107            & \url{https://www.savemyexams.com/}                     \\
ResearchGate      & 3,680            & \url{https://www.researchgate.net/}                    \\
Quizlet           & 301              & \url{https://quizlet.com/}                             \\
Study             & 1,294            & \url{https://study.com/}                               \\
\textbf{Total}    & \textbf{292,254} & \textbf{(We clean out and sample 8,000 of them)} \\ \bottomrule
\end{tabular}
}
\caption{The statistics of the Geoscience QA data collection.}
\label{tab:geoqa}
\vspace{-1em}
\end{table}

Aiming to make good use of these signals, we re-structure the data into \textit{<input, output>} pairs for tuning on tasks of \textit{Explanation, Named Entity Recognition, Reasoning, Fact Verification, Summarization, Text Classification, Word Semantics, and Question Answering}. For better understanding, all the scripts will be open-sourced after the final draft, and the details are illustrated as follows:

{\bf Explanation.} To construct the data used for training the skills of word explanation, we digitize the geoscience dictionaries, taking all the words and their explanations inside the dictionaries (Signal G7). Moreover, we also include the related entries of geography in Wikipedia to construct the dataset for the explanation tasks.

{\bf Named Entity Recognition.} Refer to Signal G5, GAKG preserves a connection between papers and knowledge entities. These entities are extracted from abstracts. Meanwhile, Wikipedia utilizes hyperlinks between Wikipedia pages. Consequently, we use the paragraphs (abstracts of papers and pages in Wikipedia) as inputs and the mentions (key entities in papers and mentions in Wikipedia pages) as outputs and re-structure the data to input-output pairs.

{\bf Reasoning.} According to Signal G6, GSO contains many relations between geoscience knowledge entities (or called concepts), one of which is the co-occurrence relation, i.e., two geoscience knowledge entities co-occur in the same paragraph according to~\cite{Xu2023ExploringAV}. We take such co-occurrence entities along with their corresponding paragraphs as input and the relation-existence as the output to model how an idea comes up. Specifically, we curate samples of co-occurrence concepts in geoscience themes, which originate from highly-cited geoscience papers using Signal G2, G3, and G7. We invite geoscientists to verify the relation-existence for each pair of co-occurrence concepts. In this way, we can endow the model with the ability to reason for ideas generation.

{\bf Fact Verification.} As for Signal G8, we collect the data from Wikipedia and papers in geoscience, and we re-structure the explicit declarative sentences. If we take a sentence with an opposite meaning as input, the output would be "False", while we handle the original sentence as input, the output would be "True".

{\bf Summarization.} We re-structure the Signal G1 and G9 for summarization task. In academic papers, a paper's abstract is the summary of the full text of the article, and the title is a further summary of the abstract. Meanwhile, the one-sentence summary of papers often exists in the related works in papers. These pieces of information provide us with supervised pairs. Moreover, Signal G3 provides us with illustrative references for papers, a kind of supervised data for summary. In this way, we can get the text summary of the training data. 

{\bf Text Classification.} Based on Signal G2, we have 18 disciplines for each knowledge point in the DDE platform, and geoscience dictionaries have eight fields to classify each term. Consequently, we can construct the dataset for text classification. 

{\bf Word Semantics.} As mentioned in Signal G8, GSO monitor the semantic relationship between geoscience-related knowledge points. In this task, we will train the model by asking about geoscience-related entities' synonyms and hyponymy relations and answering with related entities.

\begin{figure}[!t]
    \centering
    \includegraphics[width=\linewidth]{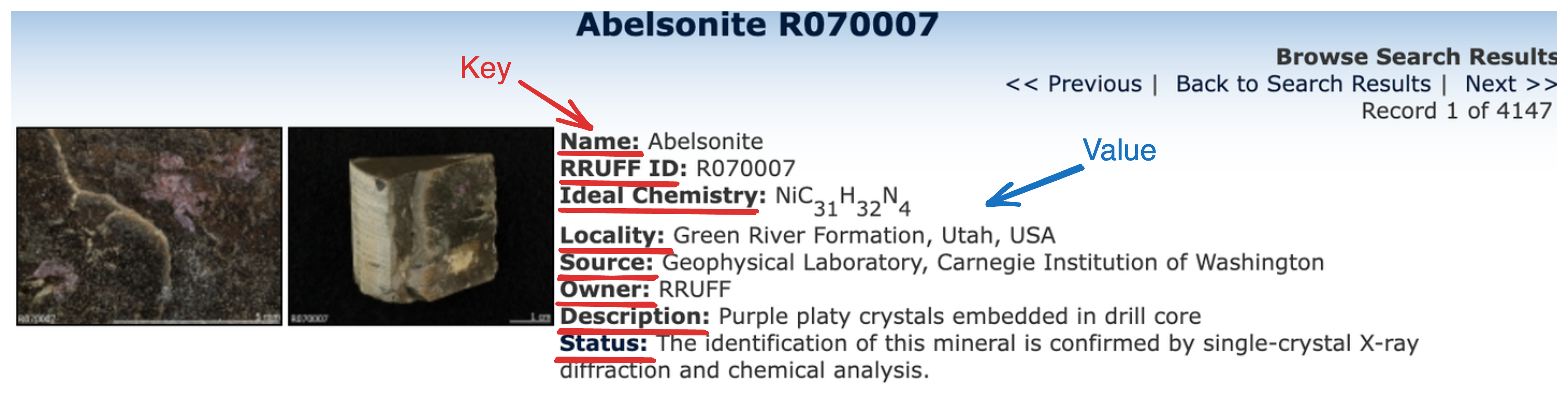}
    \caption{An example for GeoSignal re-structure when processsing the geoscience website \href{https://rruff.info/}{RRUFF}.}
    \label{fig:rruff}
    % \vspace{-1em}
\end{figure}

{\bf Question Answering.} As mentioned in G10, factual knowledge is the most essential part of model training. We re-structure four kinds of supervised data as follows:
\begin{itemize}[leftmargin=0.8em]
    \item We clean the question-answer pairs inside the Geoscience QA ensemble.
    \item We combine key-value pairs on the structured geoscience domain websites (the first six records in~\autoref{tab:geoqa}). Take the page of R070007\footnote{\url{https://rruff.info/all/display=default/R070007}} in RRUFF as an example, which is about the information of a sample of Abelsonite. We use the key of this table (shown in \autoref{fig:rruff}) as the question and the value as the answer to re-structure QA question-answer pairs.
    \item We make ChatGPT ask and answer questions to itself through Self-instruct methods and further invite experts in geoscience to validate them, ensuring the correctness of the data.
    \item Signal G4 provides us with illustrative references towards illustration and table in paper, and we combine the special token to form a simple question dataset to train the model learning and further explain the academic content.
\end{itemize}

It is worth mentioning that when we re-structure the GeoSignal data involving citations, graphs, and tables, we add the special token used in the pre-training to distinguish the unique data distribution. So we can maintain a strict format specification and consistency with further pre-trained data.

After the procedure above, we obtain a large number of supervised data, and we sample and clean the data to build the instruction tuning data GeoSignal since the data quality is much more important than quantity. We will open-source the pre-processing scripts. The statistics of GeoSignal are listed as~\autoref{tab:geosignal_task}.

% \usepackage{booktabs}
\begin{table}[!t]
\begin{tabular}{@{}llr@{}}
\toprule
\textbf{Tasks}               & \textbf{Records} & \textbf{Total (Cleaned)} \\ \midrule
Named Entity Recognition     &     6,252,268      &    2,400      \\
Reasoning                    &    1,200        &    600         \\
Fact Verification            &    168,424      &     8,000      \\
Summarization                &    3,279,336        &   800        \\
Text Classification          &    8,313      &        2,000    \\
Word Semantics               &    826,194         &   6,400         \\
Explanation                  &  731,374        &     4,200   \\ 
Question Answering           &    11,360,163    &     15,349     \\
\midrule
\textbf{Entire GeoSignal}    & \textbf{22,627,272}  & \textbf{39,749}      \\ \bottomrule
\end{tabular}
\caption{The statistics of GeoSignal are categorized by tasks.}
\label{tab:geosignal_task}
% \vspace{-2em}
\end{table}

\subsection{Evaluation on Expertise in Geoscience: GeoBench}

Lastly, in order to evaluate the language models for solving geoscience questions and the capacity to understand and utilize the geoscience knowledge, we extract the data from various Question-answer websites, crawl several open-source test websites, and finally construct a benchmark, named \textbf{GeoBench}.

\paragraph{\textbf{NPEE}} First, we collected National Postgraduate Entrance Exam questions on geology and geography in the past five years. We chose the text-only questions and translated them into English since the base model is LLaMA. We invited a professional translator who specialized in geoscience-related works to translate the questions and corresponding answers and finally obtain \emph{182} multiple-choice questions, \emph{150} fill-in-the-blank questions, \emph{454} word-explanation tasks, and \emph{335} essay questions. Since the fill-in-the-blank questions, word-explanation tasks, and essay questions are hard to evaluate, we make them subjective tasks, while the multiple-choice questions are objective.

\paragraph{\textbf{APTest}} we also collect AP (Advanced Placement) examinations, which are exams offered in the US by the College Board and are taken each May by students. We collect and clean \emph{1,395} multiple-choice questions about geology, geography, and environmental science. 

\begin{table}[!t]
\resizebox{\linewidth}{!}{%
\begin{tabular}{@{}l|ll@{}}
\toprule
\textbf{Tasks}                                    & \multicolumn{2}{l}{\textbf{Question Sample (with prompt)}} \\ \midrule
                                                  & Question & \begin{tabular}[c]{@{}l@{}}The interface between crust and mantle is called:\\ Choose from:\\ A. Gutenberg\\ B. Conrad\\ C. Moho\\ The answer is:\end{tabular}  \\
\multirow{-2}{*}{\textbf{Objective Question}}        & Answer   & {\color[HTML]{333333} C. Moho} \\ \midrule
                                                  & Question & What is Translational fault in geoscience? \\
\multirow{-2}{*}{\textbf{Subjective Question}}    & Answer   & \begin{tabular}[c]{@{}l@{}}The two walls are relatively staggered along \\ the strike direction of the fault plane.\end{tabular}                                                                                                       \\ \bottomrule
\end{tabular}%
}
\caption{Ground truth samples in GeoBench.}
\label{tab:bsample}
\end{table}

To sum up, There are \emph{183} multiple-choice questions in NPEE and \emph{1,395} in total in the AP Test, constituting the objective task set. Meanwhile, we gather all \emph{939} subjective questions in NPEE to be the subjective tasks set and use \emph{50} to measure the baselines with human evaluation. In the experiment sessions, we further discuss the evaluation metrics on these tasks and give the example of GeoBench in~\autoref{tab:bsample}.

%% file: train.tex
\section{Training the \geolm{}}

In this section, we establish a recipe for tuning a large language model on a specific domain and share the settings we adopt to train the \geolm{}.

\subsection{Geoscience Domain Adaptation Recipe}

Since geoscience is a relatively secondary or arcane field of study, there are few language models for such scenarios. However, advanced natural language models and tools can help geoscientists with data mining and knowledge discovery in their research fields. Therefore, learning a language model for knowledge understanding, summary, and QA is necessary. Meanwhile, geoscience has a rich knowledge accumulation, such as academic papers and scientific reports, which has established a data foundation for training large-scale language models. Consequently, Based on the data in the field of geosciences, we explored a recipe for scientific domain adaptation and finally obtained \geolm{}.

\begin{figure}[!t]
    \centering
    \includegraphics[width=\linewidth]{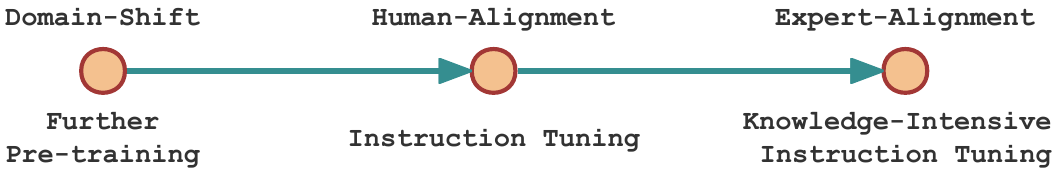}
    \caption{Training recipe for domain large language models.}
    % \jh{please change name of the figure to be consistent with others}}
    \label{fig:recipe}
    \vspace{-1em}
\end{figure}

As shown in~\autoref{fig:recipe}, scientific domain adaptation has three main steps. First, we use domain-specific text corpus to further pre-train the base model. In this paper, we use LLaMA as the base model. Second, since instruction tuning can make the language models generate content following human instructions, we can first do instruction tuning with general instruction-tuning data, such as Alpaca, and natural instruction. Lastly, after learning the paradigm to follow the instructions, the model can learn more information from the restructured domain knowledge, which we call expertise-instruction tuning. In the ablation experiments, we will further verify the correctness of this recipe.

\subsection{Further Pre-training}

During the stage of further pre-training on geoscience text corpus, We initialize the LLaMA-7B~\cite{Touvron2023LLaMAOA} checkpoints and train it on \emph{5.5B} tokens geoscience corpus. 

The entire parameters of LLaMA-7B (6.7B trainable parameters) are further pre-trained for one epoch on 4 NVIDIA A100-SXM-40GB GPUs, and the training takes \emph{214 hours}. In this stage, we set a learning rate of 1e-5, with a global batch size of 128 and a micro-batch size of 2. The incremental steps of the train are \emph{30,140 steps} (1,000 for warm-up). Finally, we call the model obtained after the further pre-train \textit{GeoLLaMA} for better distinction.

\subsection{Instruction tuning}

% \jh{I'd suggest to start with instruction tuning rather than PEFT. Instruction tuning is your core component here, while PEFT is just a specific technique to perform instruction tuning}

After further pre-training on geoscientific text data, we obtained a model that experienced domain shift. However, at this stage, the model could only accomplish the next token generation task that adhered to geoscientific knowledge distribution. In order to make the model compliant with human instructions, we employed multi-task training. Human instructions took various forms during this process, as our recipe describes, including general task instructions like the Alpaca and knowledge-intensive instructions like GeoSignal. Through experimentation, we discovered that first conducting general instruction learning followed by knowledge-intensive instruction learning helped enhance the performance of our model, far surpassing the results obtained from mixed training. 

During the instruction learning phase, we introduced parameter-efficient fine-tuning (PEFT) to help us achieve the mission of training in a low-resource setting. As mentioned in~\cite{Hu2021LoRALA}, the weight updates during the fine-tuning process also have a low ``intrinsic rank'' during adaptation. Therefore, according to LoRA, a hidden layer $h=W_{0}x, W_{0} \in R^{d*k}$, the modified forward pass yields:
\begin{equation}
    h = W_{0}x + \Delta Wx = W_{0}x + BAx,
\end{equation}
where $B \in R^{d*r}, A \in R^{r*k}$ and $r<<min(d,k)$ are two low ``intrinsic rank'' matrix containing trainable parameters. Moreover, after further pre-training the LLaMA, the adaptation to the field of geoscience is more comprehensive. During the instruction tuning stage, the target is to train the model to align with humans and experts. We use Low-Rank Adaption to tune the model. 

In instruction tuning, we set a learning rate 1e-4 with a global batch size of 128. As for the LoRA setup, we set lora\_r as eight while lora\_alpha as 16. We set the lora\_target\_modules as k\_proj, q\_proj, and v\_proj, based on our experimental observation. The instruction tuning via LoRA only trains \emph{6M} parameters on one single NVIDIA GeForce RTX 3090 for 23 hours. In order to make the model perform better and inject part of the geoscience knowledge in the SFT stage, we first use alpaca instruction tuning data to train GeoLLaMA, which we recognize as Human-alignment. Then, we resume from the checkpoint obtained and continue fine-tuning the model using GeoSignal for further training. Our experimental observation shows that the performance does not improve if we mix these training data. 

In addition, to fully exploit the capabilities of \geolm{}, we adapt tool learning for scientific utilization in geoscience. Just as we choose LoRA when using GeoSignal to fine-tune GeoLLaMA to align with humans and experts, we also use it when training \geolm{} to use tools. In this process, we refer to the~\cite{Qin2023ToolLLMFL} and use our tool dataset learning to search geoscience-related knowledge and literature. This makes \geolm{} an applicable foundation model that can call external API autonomously. The tool learning via LoRA only trains \emph{4M} parameters on eight NVIDIA GeForce RTX 3090 for 21 hours, with a learning rate of 1e-5 and a max content length of 2,048.

%% file: evaluation.tex
% \begin{figure*}[h]
%     \centering
%     \includegraphics[width=\linewidth]{img/output.png}
%     \caption{The output cases of the \geolm{}.}
%     \label{fig:case}
% \end{figure*}
\section{Evaluation and Results}

This section illustrates the evaluation methods and results of \geolm{} and related baselines. \textbf{GeoBench} consists of two kinds of tasks: one is subjective, and one is objective. In this part, we choose four baseline models: Galactica-6.7b~\cite{Taylor2022GalacticaAL}, MPT-7B~\cite{MosaicML2023Introducing}, Vicuna-7B~\cite{vicuna2023}, LLaMA-7B~\cite{Touvron2023LLaMAOA} and Alpaca-7b~\cite{alpaca}.

\subsection{Objective tasks in GeoBench}

For objective tasks like multiple-choice tasks (GeoBench-AP and multiple-choice in GeoBench-NPEE), we prompt appropriately, ending with the phrase ``\textit{The answer is}'', calculate the $Softmax$ of the probability of next token among the choice label (\textit{e.g., A, B, C, D, sometimes E}), and finally gain the score of the \textbf{Accuracy} based on these test ground truth.

\begin{figure}[!t]
    \centering
    \includegraphics[width=0.65\linewidth]{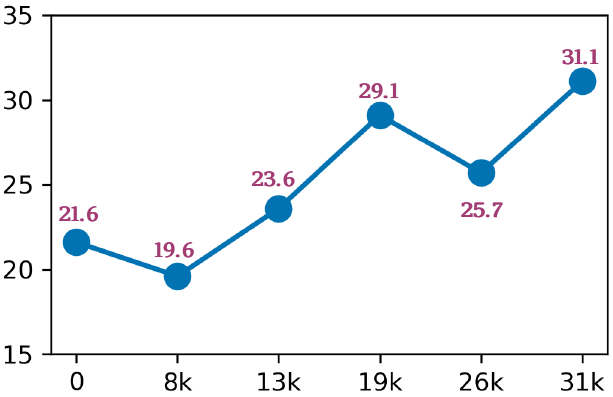}
    \caption{Accuracy scores at selected training steps of \geolm{} on the Objective tasks in GeoBench.}
    \vspace{-1em}
    \label{fig:scaling}
\end{figure}

First, we evaluate all the saved checkpoints, as shown in~\autoref{fig:scaling}. We can find that as the tokens seen by the model gradually scale up, the model's performance on our benchmark is improving. This result indicates that the model learns geoscience knowledge in further pre-train.

Moreover, compared with the baselines, shown in~\autoref{tab:baselines}, we can see that \geolm{} outperforms the model with a similar size over the NPEE dataset. However, in the AP Test, \geolm{} is similar to the Galactica model since geoscience learned in high school is human geography and environmental science, including in the training corpus of Galactica. 
% \usepackage{booktabs}
\begin{table}[!t]
\begin{tabular}{@{}lcc@{}}
\toprule
\textbf{Baselines} & \textbf{NPEE} & \textbf{APTest} \\ \midrule
Gal-6.7B           & 25.7               & \textbf{29.9}      \\
LLaMA-7B           & 21.6               & 27.6               \\
MPT-7B             & 28.4               & 26.0               \\
Vicuna-7B          & 26.4               & 16.8               \\
Alpaca-7B          & \underline{31.1}   & 29.1               \\
K2-7B (Ours)         & \textbf{39.9}      & \underline{29.3}   \\ \bottomrule
\end{tabular}
\caption{comparison among baselines on Objective tasks in GeoBench. The best number is bolded, while the second best is underlined.}
\vspace{-2em}
\label{tab:baselines}
\end{table}

\subsection{Subjective tasks in GeoBench}

For subjective tasks (mainly in GeoBench-NPEE), we use automatic methods, \textbf{GPTScore}~\cite{Fu2023GPTScoreEA} and \textbf{perplexity} to evaluate the quality of the output. 

GPTScore utilizes generative pre-trained models' emergent abilities (e.g., zero-shot instruction) to score generated texts. According to~\cite{Fu2023GPTScoreEA}, we calculate the vanilla score, which is a negative loss, with an evaluator of GPT-2~\cite{radford2019language}.  %as follows equation.
In addition, perplexity is computed with GPT-2 on the generated text and measures the fluency of the generations. 

% \begin{equation}
% \operatorname{GPTScore}(\boldsymbol{h} \mid d, a, \mathcal{S})=\sum_{t=1}^m w_t \log p\left(h_t \mid \boldsymbol{h}_{<t}, T(d, a, \mathcal{S}), \theta\right)
% \end{equation}

% where $w_{t}$ is the weight of the token at position $t$. T(·) is a prompt template that defines the evaluation protocol, which is usually task-dependent and specified manually through prompt engineering.

% \jh{which model is used to score? ChatGPT? (if so, plz also specify which version of ChatGPT.)}

Furthermore, referring to geoscientists, we collect 50 open geoscience questions and gather ten geoscience research practitioners to evaluate the output of baseline models. We evaluate the models on three metrics: 1) rationality, whether the generated content of the model is technical rationality or not; 2) correctness, whether the content generated by the model is reliable or not; 3) consistency, whether the generated content always stays in the topic. All the scores scale from 1 (poor) to 3 (good), with 2 indicating acceptable content. The complete results of the subjective tasks are in~\autoref{tab:subeval}.
\begin{table}[!t]
\resizebox{\linewidth}{!}{%
\begin{tabular}{@{}ccc|ccc@{}}
\toprule
\multirow{2}{*}{\textbf{Baselines}} &
\multicolumn{2}{c|}{\textbf{Automatic Evaluation}} &
\multicolumn{3}{c}{\textbf{Human Evaluation}} \\ \cmidrule(l){2-6} 
&
\multicolumn{1}{c}{\textbf{Perplexity $\downarrow$}} &
    \multicolumn{1}{c|}{\textbf{GPTScore $\uparrow$}} &
\multicolumn{1}{c}{\textbf{rationality}} &
\multicolumn{1}{c}{\textbf{correctness}} &
\multicolumn{1}{c}{\textbf{consistency}} \\ \midrule
\textbf{Gal-6.7B}    & 34.57           & -2.3598            & 1.96              & 1.74              & 1.79  \\
\textbf{LLaMA-7B}    & 40.07           & -1.9531            & \underline{2.24}  & \underline{2.04}  & 2.01  \\
\textbf{GeoLLaMA-7B} & \textbf{32.32} & \textbf{-1.9457}   & 2.15              & 1.89              & 2.03  \\
\textbf{Alpaca-7B}   & 40.07           & -1.9536            & 2.09              & 1.93              & \textbf{2.34} \\
\textbf{K2-7B (Ours)}& \textbf{32.32}  & \underline{-1.9487}& \textbf{2.38}     & \textbf{2.13}     & \underline{2.14} \\ \bottomrule
\end{tabular}
}
\caption{Comparsion on subjective tasks in GeoBench. The best number is bolded, while the second best is underlined.}%\jh{Why is GPTScore a negative number? Maybe better clarify how GPTScore is computed.}}
% \vspace{-2em}
\label{tab:subeval}
\end{table}

As we can see, \geolm{} performs better on rationality and correctness. At the same time, consistency stays competitive. The results indicate that our model better understands geoscience and can utilize scientific knowledge. 

% Finally, we share some examples of \geolm{} (refer to~\autoref{fig:case}) to show the capability of geoscience knowledge understanding and utilization.

\subsection{Ablation on Expert-Alignment}
% \usepackage{booktabs}
\begin{table}[!t]
\resizebox{\linewidth}{!}{%
\begin{tabular}{@{}lcc@{}}
\toprule
\textbf{Model}                 & \textbf{NPEE} & \textbf{APTest} \\ \midrule
GeoLLaMA $\rightarrow$ Dolly                 & 27.0                   & 26.3             \\
GeoLLaMA $\rightarrow$ Alpaca-GPT4           & 34.4                 & 26.5             \\
GeoLLaMA $\rightarrow$ GeoSignal             & \underline{37.2}     & \underline{27.4} \\
GeoLLaMA $\rightarrow$ GeoSignal mix Alpaca-GPT4  & 33.8                 & 23.4             \\
GeoLLaMA $\rightarrow$ Alpaca-GPT4 $\rightarrow$ GeoSignal (\geolm{}) & \textbf{39.9}   & \textbf{29.8} \\ \bottomrule
\end{tabular}
}
\caption{Results when using different SFT data. The best number is bolded, while the second best is underlined.}
\vspace{-5pt}
\label{tab:ablationonsft}
\end{table} 
To better understand the recipe for aligning the model with humans and experts, we deploy the ablation experiments to explore the detail. We treat the data constructed by self-instruct or human-annotated in the general domain or for dialogue generation as human-alignment data. At the same time, view the data annotated by experts in specific domains as expert-alignment data.
As shown in~\autoref{tab:ablationonsft}, using task-special data, such as dolly-15k, fails to achieve a good performance, while using self-instruct data, such as Alpaca-GPT4, is still not as effective as using knowledge-intensive data.
Surprisingly, we have discovered that the results are unsatisfactory if we mix the knowledge-intensive data GeoSignal with human-alignment data Alpaca. It is better to use Alpaca to align the model to follow human instruction and then use the GeoSignal to align with the experts. Moreover, LoRA is deployed only on attention layers.

\subsection{Exploration on Tool Learning}
According to the ablation experiment of Expert-Alignment, we found that the knowledge injection of large language models mainly occurs in the further pre-train stage to a certain extent, while learning to express the knowledge and generating patterns occur in the SFT stage. For different types of SFT data, the model has a learning recipe. Similar to human beings, we learn to understand words before we can read and learn to communicate before we can discuss a specific field. So, we hypothesize that there is a similar phenomenon in learning to use tools. Thus, we conducted ablation experiments to evaluate different foundation models trained with the same tools.

\begin{table}[!t]
\resizebox{\linewidth}{!}{%
\begin{tabular}{@{}lcccc@{}}
\toprule
\textbf{Model} & \textbf{rationality} & \textbf{correctness} & \textbf{consistency} \\ \midrule
Alpaca-7B $\rightarrow$ GeoTool & 1.23 & 2.45 & \textbf{2.41} \\
\geolm{}-7B $\rightarrow$ GeoTool  & \textbf{2.67} & \textbf{2.56} & 2.24 \\
\bottomrule
\end{tabular}
}
\caption{Results of evaluating the generated thought quality from foundation models while doing tool learning (scores are rated by prompting ChatGPT).}
\vspace{-5pt}
\label{tab:ablationontool}
\end{table} 

In this experiment, we perform tool instruction tuning with the GeoTool dataset on both Alpaca-7B (i.e., LLaMA model tuned on the Alpaca-GPT4 dataset) and \geolm{}-7B, which undergoes both further pre-training and instruction tuning. This comparison is similar to a student educated in a general subject and a student specializing in geosciences using a geoscience literature search tool for knowledge queries and question answering. In this work, we need to compare the generative thoughts (illustrated in~\autoref{fig:aide}). We take ChatGPT as a referee to decide which model generates better thoughts and ask ChatGPT to give a score based on the meaning of these three factors via simple prompts, following previous studies~\cite{Gilardi2023ChatGPTOC}. During this process, we use the 50 open geoscience questions mentioned in the subjective tasks above. 

Finally, according to the results shown in~\autoref{tab:ablationontool}, \geolm{} may have better thoughts on how to write search queries since \geolm{} knows more geoscience knowledge than Alpaca-GPT4 leading to a better understanding towards geoscience questions.

%% file: discussion.tex
\section{Application}

In this section, will discuss the potential application of \geolm{} in brief. We will show the use cases of \geolm{} for research assistance and knowledge reasoning.

\begin{figure}[!t]
    \centering
    \includegraphics[width=0.8\linewidth]{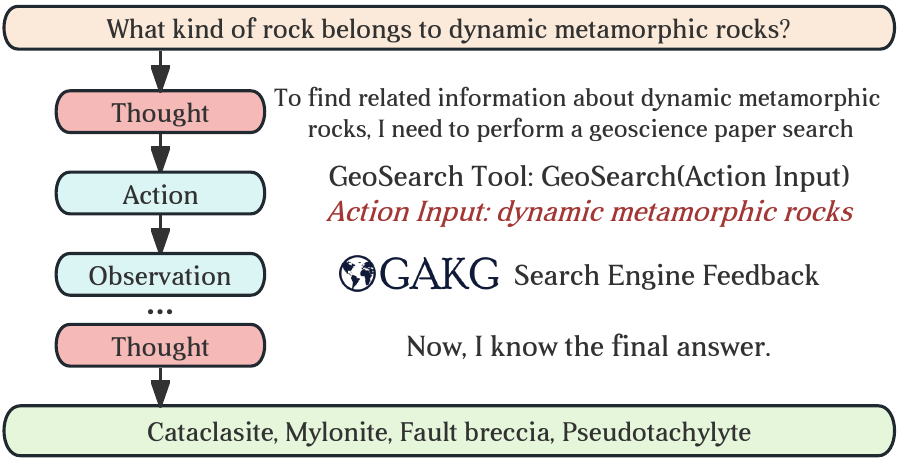}
    \caption{An example of tool augmented \geolm{} for geoscience Q\&A.}
    \label{fig:aide}
    \vspace{-1em}
\end{figure}

\subsection{Case Study 1: \geolm{} for research assistance}

As a language model, \geolm{} can understand geoscience materials and modify the statement about geoscience with suitable prompts. Since \geolm{} is a generative language model, it can generate paragraphs and statements on word description and answer generation based on the given questions. In this way, \geolm{} acts like a knowledge base and provides the geoscientist with a professional assistant.

Moreover, with the tools Augment \geolm{}, we can use external information and functions to generate reliable and promising results. Take GeoSearch as an example. After giving a question to \geolm{}, \geolm{} will give out a thought to call the API of searching on the GAKG, then update the action and generate a suitable prompt based on observation. Finally, the model can generate a thought of generating appropriate answers. \autoref{fig:aide} shows an example.

\subsection{Case Study 2: \geolm{} for knowledge reasoning}

\begin{table}[!t]
\resizebox{\linewidth}{!}{%
\begin{tabular}{l|l}
\toprule
\textbf{Concepts} & yingcheng formation, kamchatka \\ \midrule
Prompt   & \begin{tabular}[c]{@{}l@{}}Given two academic concepts:\\ Concept 1: yingcheng formation\textbackslash{}nConcept 2: kamchatka\\ Assuming you are a researcher: \\ What is the meaning of these two concepts?\\ What new knowledge can they merge to generate? \\ Please write a paper abstract to combine (fuse) these two concepts:\end{tabular} \\ \hline
Output   & \begin{tabular}[c]{@{}l@{}}Abstract: \textbf{The Kamchatka Peninula} is located in the northwestern Pacific \\ Ocean, between the Bering Sea and the Sea of Okhotsk. The region is \\ characterized by a complex tectonic setting, with the Pacific Plate sub\\ -ducting beneath the North American Plates.\textbackslash{}n\textbf{Yingcheng formation}, which\\ is a Late Cretaceous to Early Paleogene sedimentary sequence. It is compos\\ -ed of sandstone, siltstone and mudstone. These sediments were deposited \\ in a shallow marine environment.\textbackslash{}n\textbf{The Yincheng sedimeants} were formed in \\ an outer shelf environment, where the water depth was less than 500 meters. \\ During the Late Paleocene, a large-scale volcanic eruption occurred in this \\ region, resulting in volcano-sedimentation. A large amount of volcniclastic \\ sediemants, such as tuff, tufite and volcamic breccia, were interbedded with \\ sedimetary rocks.The volcaiclastics were derived from the volacnic eruptive \\ centers in eastern \textbf{Kamachatsky peninsular}.\end{tabular} \\ \hline

\end{tabular}%
}
\caption{An example for \geolm{} used for knowledge reasoning and idea generation.}
\label{tab:aide}
\vspace{-2em}
\end{table}

\geolm{} shows the potential of adapting language models to a scientific field with domain barriers. Based on the observation of the performance on subjective tasks, in this paper, we adopt \geolm{} to do the task of idea generation and knowledge reasoning in the field of geoscience to find the potential scientific relations, which will be brought to the fore. Consequently, \geolm{} can generate new ideas in the form of abstracts according to the co-occurrence concepts mentioned above. \autoref{tab:aide} shows an example.

%% file: conclusion.tex
\section{Conclusion}
% \jh{usually the discussion and conclusion are in the same section -- you don't need the conclusion section but shortly conclude in the discussion section. Currently the discusssion + conclusion sections are too long, should be compressed. for example, in the submission you don't need to write too much on future works.}
% In this session, we will discuss the limitation, potential applications, and future works of \geolm{}, based on our attempt and observation of large language model techniques applied to science fields like geoscience.
In this paper, we introduce \geolm{}, the first-ever large language model and foundation model in the geoscience field. \geolm{} can answer geoscience questions and follow geoscientists' instructions with its geoscience professionalism. We construct the first geoscience-supervised instruction data, GeoSignal. Meanwhile, we build GeoBench, the first NLP benchmark in geoscience to evaluate the capability on geoscience knowledge understanding and utilization.
On the geoscience benchmarks collected, \geolm{} shows its professionalism and effectiveness compared with other similar-size language models. 
Moreover, we share the \geolm{} potential applications including knowledge reasoning and research assistant.
Finally, we open-source all the code, data and \geolm{} model weights at \url{https://github.com/davendw49/k2}.